\documentclass[a4paper,fleqn]{cas-dc}
\pdfoutput =1

\usepackage[authoryear,longnamesfirst]{natbib}
\usepackage{threeparttable}

\def\tsc#1{\csdef{#1}{\textsc{\lowercase{#1}}\xspace}}
\tsc{WGM}
\tsc{QE}

\begin{document}
\let\WriteBookmarks\relax
\def\floatpagepagefraction{1}
\def\textpagefraction{.001}
\let\printorcid\relax

\shorttitle{SpikeSEE: An Energy-Efficient Dynamic Scenes Processing Framework for Retinal Prostheses}

\shortauthors{Chuanqing Wang, Chaoming Fang, Yong Zou, Jie Yang, Mohamad Sawan}  

\title [mode = title]{SpikeSEE: An Energy-Efficient Dynamic Scenes Processing Framework for Retinal Prostheses}  



\author[1]{Chuanqing Wang}
\author[1]{Chaoming Fang}
\author[2]{Yong Zou}
\author[1]{Jie Yang}
\fnmark[*]
\author[1]{Mohamad Sawan}
\fnmark[*]

\affiliation[1]{organization={Center of Excellence in Biomedical Research on Advanced
Integrated-on-chips Neurotechnologies, School of Engineering, Westlake university},
            city={Hangzhou},
            postcode={310024}, 
            state={Zhejiang},
            country={China}}

\affiliation[2]{organization={Beijing Institute of Radiation Medicine},
            city={Beijing},
            postcode={100850},
            country={China}}

\cortext[1]{Corresponding authors at: Center of Excellence in Biomedical Research on Advanced
Integrated-on-chips Neurotechnologies, School of Engineering, Westlake university, Hangzhou, 310024, China. (E-mail address: yangjie@westlake.edu.cn, sawan@westlake.edu.cn)}

\begin{abstract}
Intelligent and low-power retinal prostheses are highly demanded in this era, where wearable and implantable devices are used for numerous healthcare applications. In this paper, we propose an energy-efficient dynamic scenes processing framework (SpikeSEE) that combines a spike representation encoding technique and a bio-inspired spiking recurrent neural network (SRNN) model to achieve intelligent processing and extreme low-power computation for retinal prostheses. The spike representation encoding technique could interpret dynamic scenes with sparse spike trains, decreasing the data volume. The SRNN model, inspired by the human retina's special structure and spike processing method, is adopted to predict the response of ganglion cells to dynamic scenes. Experimental results show that the Pearson correlation coefficient of the proposed SRNN model achieves 0.93, which outperforms the state-of-the-art processing framework for retinal prostheses. Thanks to the spike representation and SRNN processing, the model can extract visual features in a multiplication-free fashion. The framework achieves 12 times power reduction compared with the convolutional recurrent neural network (CRNN) processing-based framework. Our proposed SpikeSEE predicts the response of ganglion cells more accurately with lower energy consumption, which alleviates the precision and power issues of retinal prostheses and provides a potential solution for wearable or implantable prostheses.

\end{abstract}




\maketitle

\section{Introduction}

Retinal prosthesis is a promising medical device to restore vision for patients with severe age-related macular degeneration (AMD) or retinitis pigmentosa (RP) (\cite{palanker2022simultaneous, busskamp2010genetic}). These two diseases are caused by the apoptosis of photoreceptors, but ganglion cells remain intact and functional (\cite{den1999mutations}). Retinal prostheses, including a vision sensor, processing framework, and tissues stimulator, can mimic the behavior of degenerative cell layers and make good use of intact ganglion cells (\cite{soltan2018head}). The image sensors first perceive external scenes, and the recorded signals will be transmitted to a processing framework that replaces the function of the bypassed cell layer to generate a stimulation pattern. This device utilizes these patterns to stimulate ganglion cells to fire spikes, which can be interpreted by the human brain to produce visual perception (\cite{sawan2021emerging}).

The processing framework is the most critical part that affects the performance of the prostheses (\cite{turner2019stimulus, yu2020toward}). Firstly, the framework should effectively mimic the function of the bypassed layers and predict the response of multiple ganglion cells. If this framework is successful, a prosthesis can restore vision for the blind. Secondly, considering that a prosthesis works as an implantable or wearable device, it needs to serve the patient for a prolonged time. Consequently, this processing framework should be energy efficient.

There are two types of processing frameworks designed for retinal prostheses: biophysical processing-based framework and deep learning processing-based framework (\cite{meyer2017models, yu2020toward}). The biophysical processing-based framework utilizes a bio-inspired architecture to build the mapping relationship between stimulus and response (\cite{lawlor2018linear, mcfarland2013inferring}). Among several traditional studies in this field, the conductance-based neural encoding framework is the latest and most powerful one of the biophysical processing-based frameworks (\cite{latimer2019inferring}). It adopts bio-inspired excitatory and inhibitory filters to process input stimulus, whose outputs are rectified with a conductance transmission mechanism. After processing with nonlinear function and spike generation process, the final predicted patterns are obtained. This framework achieves state-of-the-art performance in biophysical processing-based frameworks. However, it can only model the response of one ganglion cell, which is not suitable for retinal prostheses that need to mimic the behavior of multiple cells.

On the other hand, the deep learning processing-based framework, as an alternative solution for retinal prostheses, can simultaneously predict the response of multiple ganglion cells (\cite{lozano20183d, yan2020revealing}). Recently, the convolutional recurrent neural network (CRNN) processing-based framework outperforms other frameworks in modeling the response of dynamic scenes (\cite{zheng2021unraveling}). This framework combines the convolution and recurrent layers to improve the spatial and temporal feature extraction ability, which is suitable for processing continuous spatiotemporal signals. 
However, this framework utilizes a CMOS image sensor to record external scenes with a series of frames, generating high data redundancy in the perception end. Moreover, it adopts floating-point multiplication in the processing component, resulting in a large amount of data processing. The frame-based recording technique and processing method of the CRNN processing-based framework require a high-energy level, making it unsuitable for retinal prostheses. In addition, this framework utilizes two-convolution layers and one-recurrent layer, whose structure and floating-point multiplication operation lack biological similarity (\cite{wang2022neurosee}).

To solve the bottleneck in CRNN processing-based framework and satisfy the requirement of the processing framework for retinal prostheses, we propose an energy-efficient dynamic scenes processing framework (SpikeSEE) to mimic the response of multiple ganglion cells. This framework utilizes a dynamic vision sensor to encode external video information into sparse spike trains. This encoding technique can decrease power consumption during the perception and transmission between a sensor and processing framework. Besides, the framework adopts a bio-inspired spike recurrent neural network (SRNN) model to handle prediction tasks. Inspired by the human retina consisting of three excitatory cell layers and two inhibitory cell layers, this model adopts three spike layers and two recurrent blocks to predict the response of multiple ganglion cells. Its performance outperforms the fitness accuracy of the convolutional recurrent neural network (CRNN) model. In addition, the spike processing method of the framework is helpful in decreasing the power consumption without any floating-point multiplication. The proposed SpikeSEE with better prediction accuracy and energy efficiency is a promising processing framework to bring more convenience to the blind.

The remaining parts of this paper are organized as follows. In the section \uppercase\expandafter{\romannumeral2}, we introduce the development process of biophysical processing-based framework and the deep learning-based framework, and summarize the challenges in this research field. The proposed processing framework including dynamic vision sensor and SRNN model is described in section \uppercase\expandafter{\romannumeral3}. Section \uppercase\expandafter{\romannumeral4} elaborates the acquisition and processing of spike signals and the training details of processing models.
The results about the prediction accuracy and energy efficiency of the framework are shown in section \uppercase\expandafter{\romannumeral5}. The last two sections focus on the limitation and future work of the processing framework for retinal prostheses.

\section{Processing Frameworks for Retinal Prostheses}

\subsection{Biophysical processing-based framework} 

Recently, neuroscientists discovered several signal processing mechanisms of retina cells and connection modes between cells in the past decades (\cite{gollisch2010eye, kong2018efficient}). Subsequently, computational scientists utilized these findings to construct biophysical processing-based frameworks for building the mapping relationship between stimulus and response (\cite{schroder2020system}). 
The receptive field characteristic of the retina was first discovered by Gilbert et al., which works as a spatiotemporal filter to process the scenes (\cite{gilbert1992receptive}). Besides, some studies found the nonlinear rectification function embodied in neurons (\cite{lawlor2018linear, farrell2021autoencoder}). Based on these findings, a linear-nonlinear Poisson framework, including a linear filter and a nonlinear function, was designed to model the response of ganglion cells. Then, some studies observed that excitatory and inhibitory cells have different processing features (\cite{gollisch2008rapid, kim2021periodic}). It inspired authors to build a multiple filter processing framework that adopted excitatory and inhibitory filters. Recently, the conductance transmission mechanism of retina cells was revealed, which became the innovation component of the conductance-based neural encoding framework (\cite{latimer2019inferring}).

A conceptual map of conductance-based neural encoding frameworks is depicted in Figure \ref{system_architecture}(a) to demonstrate the technical details of this type of framework. The framework adopts an integrated image sensor to record the external scenes with a series of frame signals. Then, a conductance-based neural encoding model utilizes recorded frame signals to mimic the firing rate of ganglion cells, composed of excitatory and inhibitory filters, conductance transmission mechanism, nonlinear function, and a recurrent filter. The predicted firing rates are used to stimulate ganglion cells. However, the perception in the visual cortex is limited because the framework can only mimic the response of one ganglion cell (\cite{meyer2017models}). If we extend this model in parallel to predict more neurons' responses, the model will become very complex and consume a huge amount of energy in the prediction process, which cannot satisfy the requirement of the processing framework for retinal prostheses. 

\subsection{Deep learning processing-based framework} 

Deep learning models have made great strides in image recognition and classification (\cite{lecun2015deep}). Its strong feature extraction ability also attracted authors to build deep learning processing-based frameworks for retinal prostheses. Firstly, a three-layer CNN processing-based framework was proposed to predict the response of ganglion cells to natural stimuli, such as scene images from ImageNet (\cite{mcintosh2016deep}). Then, an extended study adopted a pre-trained model trained in an image classification task in their proposed framework (\cite{cadena2019deep}). This framework achieved better accuracy than a directly training framework with the same architecture. Compared with the above frameworks that focus on fitting the response of static natural images, some researchers move their attention to dynamic video signals. A 3D CNN processing-based framework was designed to fit the response of ganglion cells to video signals (\cite{lozano20183d}). It was a successful attempt on this type of prediction task. Then, a CRNN processing-based framework with two convolution layers and one recurrent layer was proposed to model ganglion cells' response to videos, which intends to further improve the prediction performance in this task (\cite{zheng2021unraveling, batty2017multilayer}).

\begin{figure*}[!t]
\centering
\includegraphics[width= 0.96\textwidth]{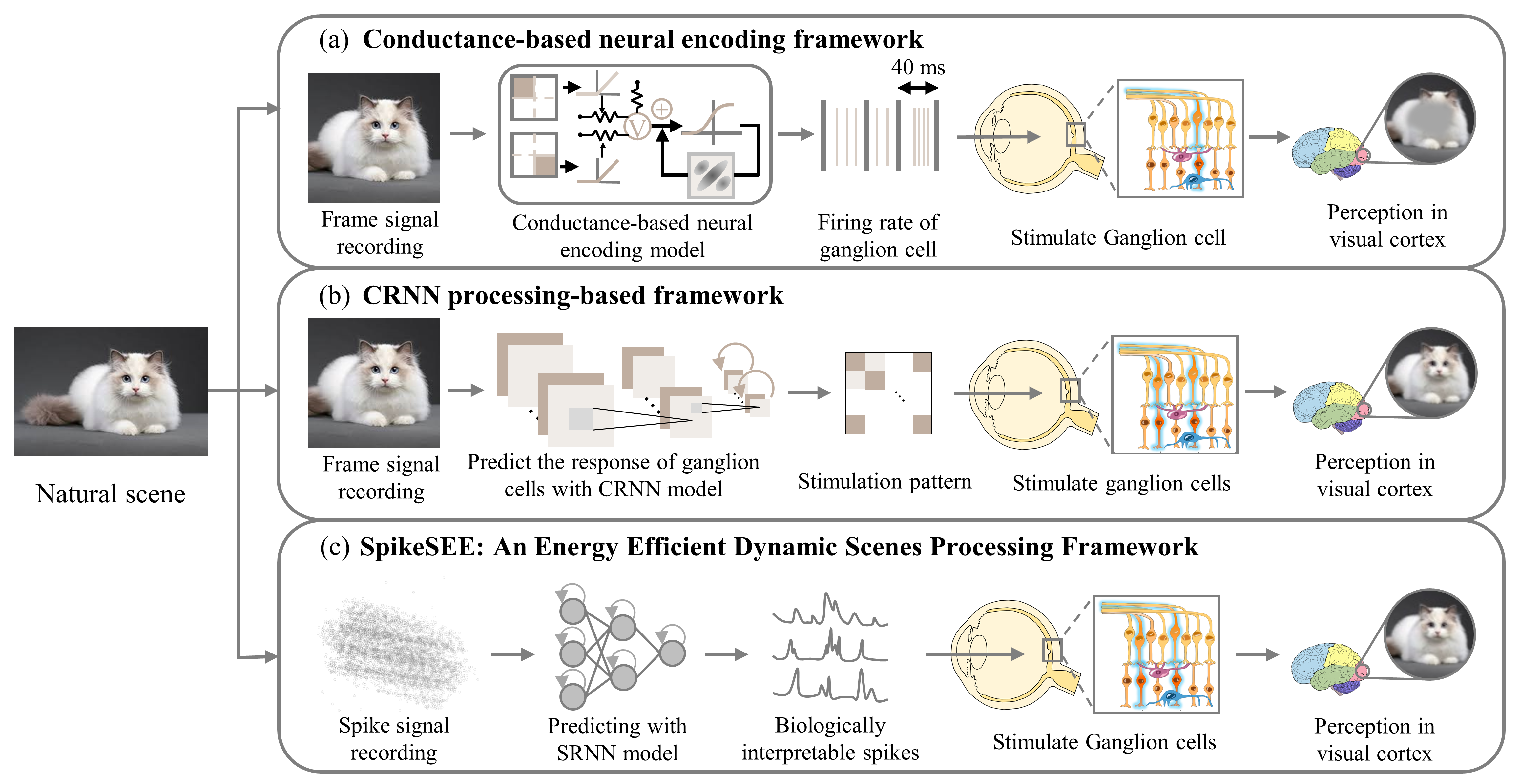}
\caption{Conceptual diagram of two existing processing frameworks and the proposed energy-efficient dynamic scenes processing framework (SpikeSEE). (a) Conductance-based neural encoding framework and CRNN processing-based framework, as the representative framework in biophysical processing-based framework and deep learning processing-based framework, are selected to show its architecture and technique details. (b) The proposed SpikeSEE adopts a spike representation encoding technique to record external scenes and utilizes the SRNN model to process recorded spike signals. The visual cortex can interpret the predicted spike train to generate scene perception.}
\label{system_architecture}
\end{figure*}

The technical details of the CRNN processing-based framework to predict the response of multiple ganglion cells are shown in Figure \ref{system_architecture}(b). The framework encodes scene signals into a series of frame signals. A CRNN model, including two-convolution layers and one-recurrent layer, is used to predict the response and generate the stimulation patterns. These patterns embodying rich spatiotemporal features can improve the perception compared with the above conductance-based neural encoding framework. 
However, this framework adopts an image sensor to record natural scenes with a series of frames, easily resulting in data redundancy. Besides, it adopts floating-point multiplication operations in the processing part, causing a large amount of data processing. Thus, CRNN processing-based framework with frame-based recording technique and processing method consumes a large power budget. These two existing processing frameworks are unsuitable for retinal prostheses as wearable or implantable devices with a high demand for prediction accuracy and power consumption. 

\section{SpikeSEE Architecture}
To overcome the bottleneck in the existing processing framework, it is necessary to learn from the working mechanism of the human retina, especially its perception and processing methods of it.
Photoreceptors in the human retina perceive external light and encode corresponding signals with nerve spikes (\cite{jackson2002photoreceptor}). Bipolar and ganglion cells process the input spikes and generate spike trains that would be transmitted to the LGN and the visual cortex by optic nerves (\cite{ruether2010pkcalpha, kondo2011identification}). The spike representation encoding technique and spike processing method are two important factors that make the retina powerful enough to perceive and process various natural scenes in low-power consumption conditions. It inspired us to propose an energy-efficient dynamic scenes processing framework (SpikeSEE) to conquer the high-power consumption limitation existing in the CRNN processing-based framework (Figure \ref{system_architecture}(c)). This framework utilizes a dynamic vision sensor to record the natural scene, and the recorded signals are represented as spike trains. An SRNN model is adopted in this framework to process the input signals and generate biologically spike sequences that be interpreted by the visual cortex to form visual perception. 
The SRNN model has three spike layers and two recurrent blocks; its spatial and temporal feature extraction ability helps it to achieve high prediction performance. In addition, the spike representation encoding technique of the dynamic vision sensor and spike processing method of the SRNN model could significantly decrease power consumption throughout the process. Thus, SpikeSEE is a promising processing framework to serve a new generation of retinal prostheses and restore vision.

\begin{figure}[!t]
\centering
\includegraphics[width= 0.48\textwidth]{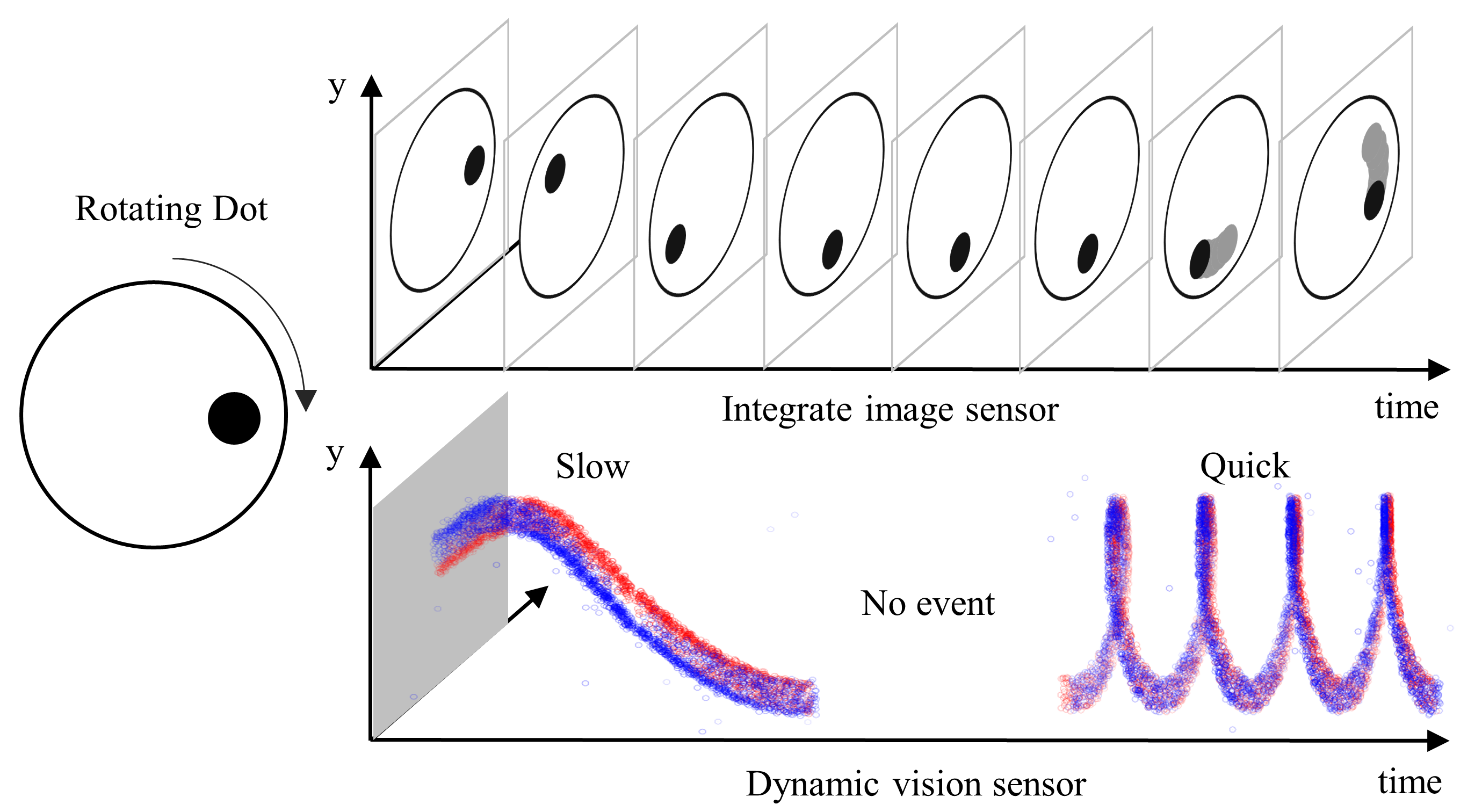}
\caption{Integrate image sensor and dynamic vision sensor are used to record a rotating dot in three different conditions. Integrate image sensor encodes the dot with a series of frame signals. It easily causes motion blur in the recorded frames for a fast rotating dot and data redundancy for a stationary dot or a slow rotating dot. The dynamic vision sensor utilizes sparse spike trains to record external information. It can provide high-quality spike sequence for a high rotating dot and decrease the power consumption by only responding to the change of the whole scene.}
\label{dvs}
\end{figure}

\begin{figure*}[!t]
\centering
\includegraphics[width= 0.98\textwidth]{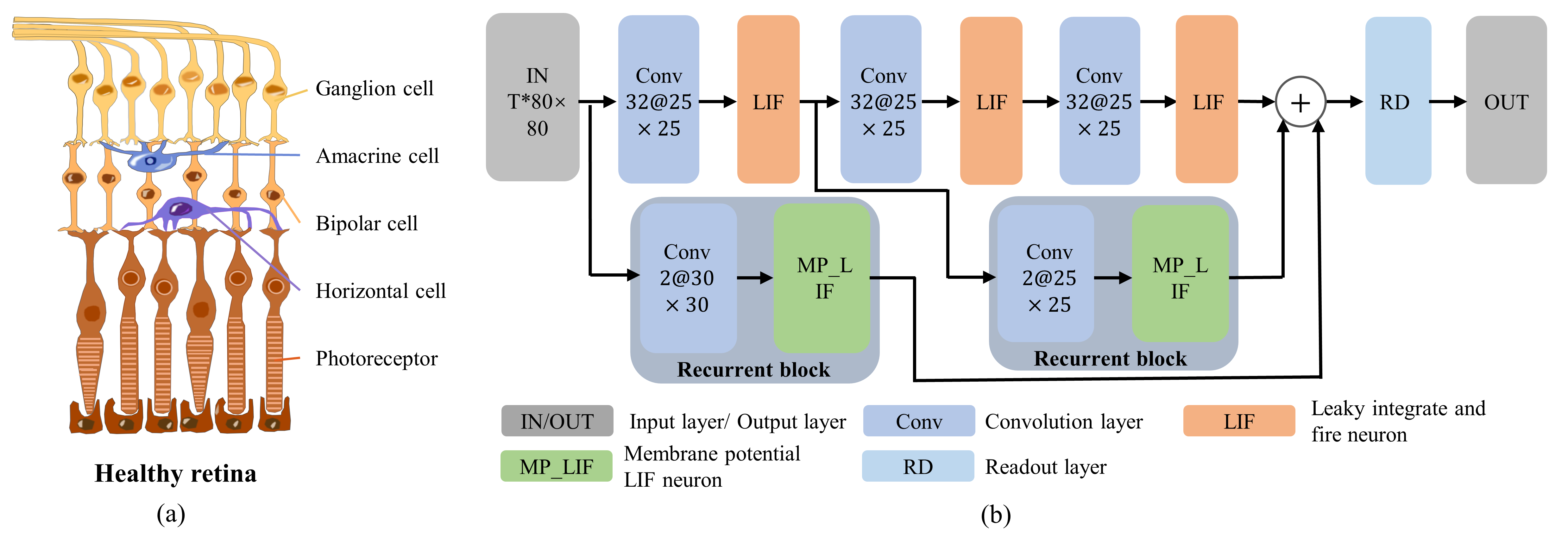}
\caption{The structure of the human retina and the proposed SRNN model. (a) The human retina is composed of three excitatory cell layers and two inhibitory cell layers. (b) The bio-inspired SRNN model adopts three spike layers and two recurrent blocks to mimic the behavior of the retina for predicting the response of ganglion cells.}
\label{srnn}
\end{figure*}

\begin{figure}[!t]
\centering
\includegraphics[width= 0.48\textwidth]{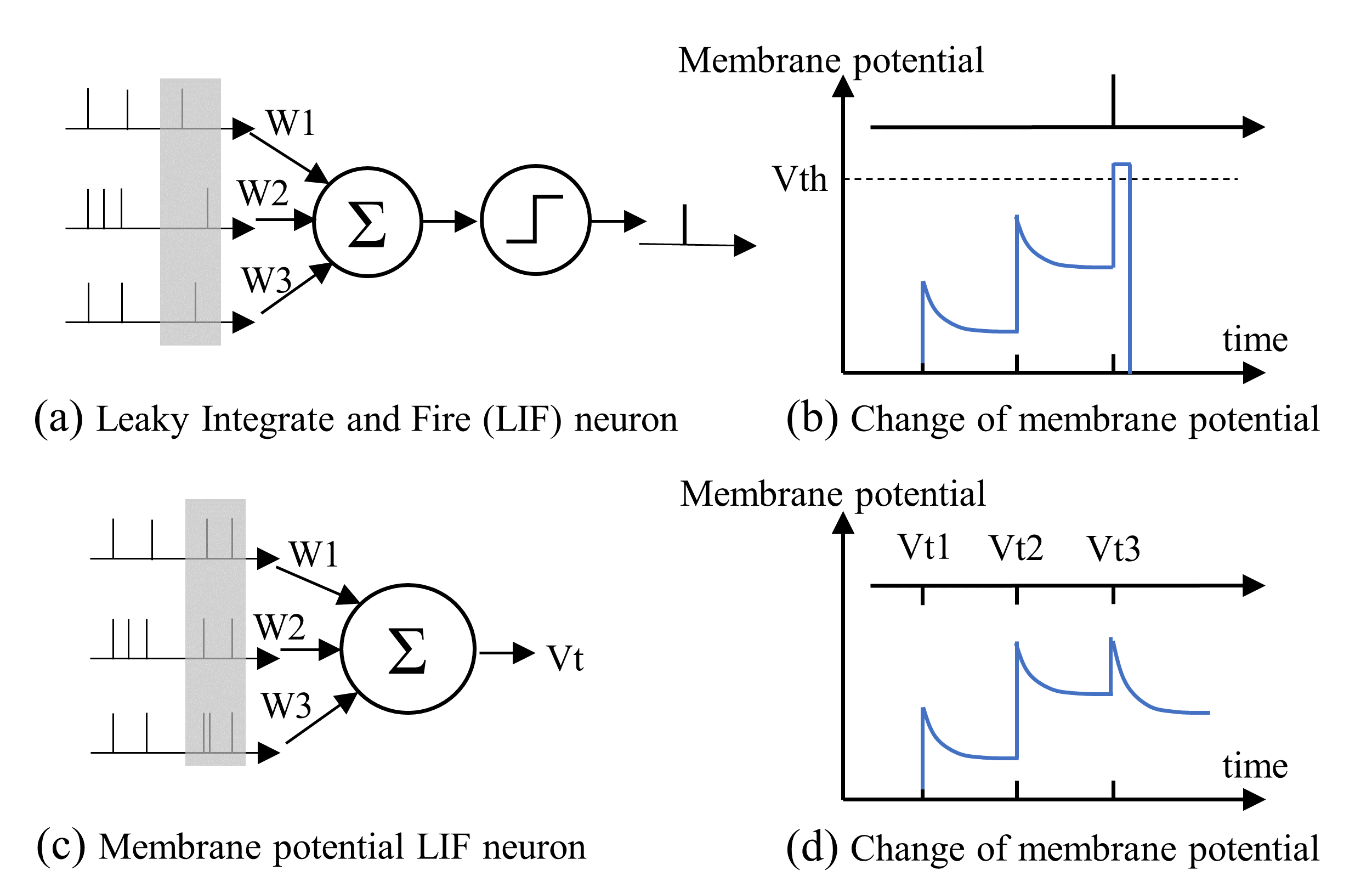}
\caption{Dynamic feature of LIF and MP\_LIF neuron: (a) working mechanism of LIF, (b) membrane potential change of LIF neuron, (c) working mechanism of MP\_LIF neuron, and (d) membrane potential change of MP\_LIF neuron. The output of this kind of neuron is the membrane potential instead of a spike.}
\label{lif}
\end{figure}

\subsection{Spike representation encoding technique}

A dynamic vision sensor (DVS) is used in our proposed processing framework to record external scenes. And the comparison between this type of sensor and a regularly integrated image sensor is shown in Fig. \ref{dvs}. In this figure, we focus on the feature of these two types of sensors in three different conditions. A fast rotating dot causes motion blur for the image sensor, while DVS with a high temporal resolution can record the whole movement completely. It could provide better input signals for the subsequent processing component. A slow rotating and stationary dot can be recorded by an integrated image sensor with a series of frames, easily leading to data redundancy. The redundant data can result in high-power consumption in the transmission and processing part. However, DVS responds to the change in the whole scene and records it with a sparse spike train. This feature helps decrease the power consumption in the perception and processing components, especially for a stationary scene. Thus, this spike representation encoding technique could provide efficient scene signals in an energy-efficient way, which is more suitable for the processing framework of retinal prostheses (\cite{ecke2021exploitation}). 

\subsection{Bio-inspired SRNN model}

The human retina, with a unique structure and spike processing method, could perceive and process external scenes, providing efficient information to the visual system for scene perception (\cite{lin2021brain}). As shown in Fig. \ref{srnn}(a), the human retina comprises three excitatory cell layers and two inhibitory cell layers. Photoreceptors, bipolar cells, and ganglion cells undertake the main signal processing task. Two other cell layers, horizontal and amacrine cells, are responsible for modulating the spike signals to improve computing efficiency, specificity, and diversity. Inspired by the unique structure of the human retina, an SRNN model with three spike layers and two recurrent blocks is proposed to mimic the behavior of the retina and improve its prediction accuracy (Fig. \ref{srnn}(b)). Three spike layers undertake the main feature extraction task, like the function of three excitatory cell layers. In addition, the recurrent block, consisting of convolution layer and membrane potential leaky integrate and fire (MP\_LIF) neuron, performs a similar function to inhibitory cell layers to make the computation efficient and diversiform.

The outstanding feature of the SRNN model is the spike processing method, which depends on the dynamic characteristics of the LIF neuron. As is shown in Fig. \ref{lif}(a), the input potential of the LIF neuron obtained by multiplying the input spikes and corresponding weights would accumulate to the existing membrane potential.
Then, the new membrane potential must compare with the threshold to determine whether to fire a spike. Fig. \ref{lif}(b) shows the LIF neuron's change in membrane potential. This neuron will generate a spike when membrane potential exceeds the threshold. Thus, LIF neurons ensure input and output signals of these spike layers are spikes rather than floating-point numbers. 
The proposed SRNN model also adopts two recurrent blocks to enhance the spatial feature extraction ability, which relies on the function of MP\_LIF neuron. The main features of MP\_LIF neuron are depicted in Fig. \ref{lif}(c). The difference between LIF neurons is that this kind of neuron does not have a comparison step, whose output is membrane potential (Fig. \ref{lif}(d)). Besides, the dynamic characteristic of MP\_LIF can be written as:

\begin{equation}
\label{eq:1}
    V_t = (1- \frac{1}{\tau}) V_{t-1} + \frac{1}{\tau} X_t 
\end{equation}
where equation (1) is similar to the function of the recurrent layer. The membrane time constant $\tau$ controls the balance between remembering $X_t$ and forgetting $V_{t-1}$ (\cite{zhu2022event}). Thus, it can be considered a simple version of the recurrent layer.

The input of three spike layers and two recurrent blocks in the SRNN model is spike signals from the dynamic vision sensor or spike layers. It can avoid the floating-point multiplication operations to decrease the power consumption. Besides, the low-firing rate in each spike layer also contributes a lot to reducing power consumption. Thus, the bio-inspired SRNN model can operate in a low power-consuming condition, which is more suitable for retinal prostheses. 

\section{EXPERIMENTAL SETUP}
To validate the performance of the proposed SpikeSEE, a completed experimental flow was drawn up that includes biological ganglion cells’ response dataset, spike signals recording technique, spike signals processing, and SRNN model. The biological ganglion cells’ response dataset is a public dataset that records the response of ganglion cells of isolated salamander retinas to dynamic videos. Then, a spike signals recording technique was used to transform dynamic video into spike event signals. And it is necessary to improve the quality of the recorded spike train with a spike-signals processing method. The optimized spike events were the basis for constructing a spike event response dataset for training the SRNN model. This model with a bio-inspired structure could be trained directly to fit the response of ganglion cells. 

\subsection{Biological ganglion cells’ response dataset}
The ganglion cells’ response dataset includes two parts, two dynamic videos and corresponding responses. These two dynamic videos are about two one-minute natural movies with a spatial resolution of 360 × 360 pixels (\cite{onken2016using}). The first scene is about a small salamander swimming in the water, and the other scene is about a tiger preying on a deer. When displaying the video to a salamander, multi-electrode arrays were used to record the response of ganglion cells. The recorded dataset includes the action potential time of 38 ganglion cells in each trial. And this recorded experiment was repeated 30 times for each video. In the training process, the average response of these 30 trials was used as the label for supervised learning. 

\begin{figure}[!t]
\centering
\includegraphics[width= 0.5\textwidth]{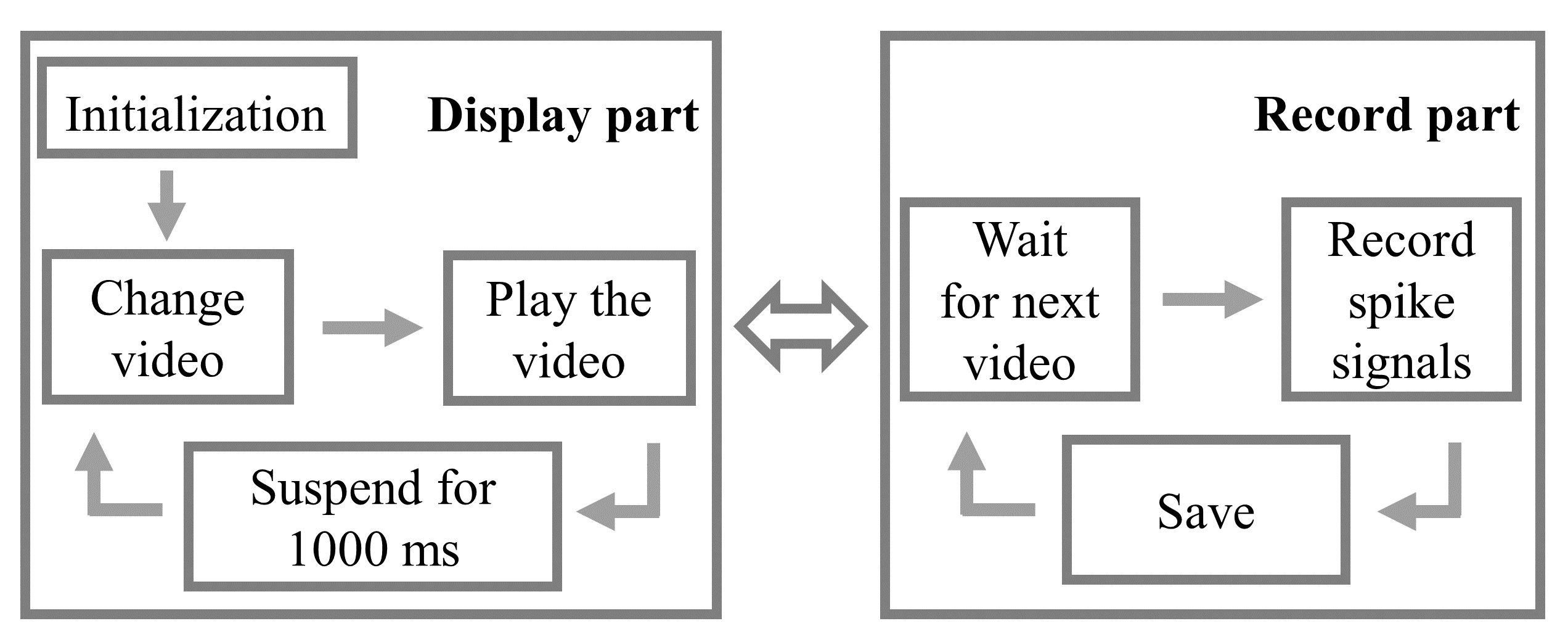}
\caption{A synchronous recording system, including display part and record part, is proposed to ensure the synchronization between playing video and collecting spike signals. }
\label{recording_system}
\end{figure}

\subsection{Spike signals recording technique}
A synchronous recording system was proposed to ensure the synchronization between playing video and collecting spike signals. As shown in Fig. \ref{recording_system}, the recording system comprises two parts: display and record parts. When the recording began, the display part needed to play the video. And the record part utilized a dynamic vision sensor to record the change in the whole scene. After spike signals acquisition of one video, the display part is suspended for 1000 ms. During this period, the recorded spike signals were saved. Then, the display part loads the other video and plays it. The dynamic vision sensor synchronously records the spike signal of this later video. The synchronization between two parts in this recording system utilized TCP/IP protocol to ensure efficient communication.

\subsection{Spike signals processing}

The recorded spike events are the basis for constructing an event dataset for training the SRNN model. There are three steps to building the dataset: denoising, deconstructing spike events, and decreasing spatial resolution. First, the recorded events were mixed with environmental noise that should be removed. A denoising method proposed by Serrano-Gotarredona was adopted to solve this problem and improve the quality of the spike train (\cite{lenero20113}). Then, it was necessary to deconstruct the spike events of one video into several spike trains, making them suitable as input data for the SRNN model. 
In this experiment, the time window parameter of each spike train was set to 33 ms, which could provide efficient features for the processing model to predict. Apart from the time window, because the spatial resolution of the dynamic vision sensor is 346 × 240 pixels, it is also unsuitable for directly processing. This study decreased the temporal resolution of spike events to 80 × 80 pixels for the SRNN model. 

\subsection{SRNN model}

As shown in Fig. \ref{srnn}(b), the proposed SRNN model comprises three-spike layers and two-recurrent blocks. The spike layer consists of a convolution layer and leaky integrate and fire (LIF) neuron. The weight parameter of the first convolution layer is Conv 32@25x25, and the parameter of the rest two-convolution layers are Conv 32@25x25 and Conv 32@25x25, respectively. The LIF neurons in these three layers are 100352, 32768, and 128, respectively. Then, the weight parameter of the convolution layer in these two-recurrent blocks is Conv 2@30x30 and Conv 2@25x25, respectively. And the number of the MP\_LIF in each recurrent block is 128. The readout layer is to bridge the output size of the third spike layer and the number of recorded ganglion cells, which is 38 in this study.

During the training process, the model input was a spike train of several time steps. The specific time step number is changed according to the experimental setup. The model output is the predicted firing rate of 38 ganglion cells in the corresponding time period. In this experiment, the Poisson loss function is adopted to calculate the model's loss, which is computed after every prediction. Then, backpropagated error passes through the LIF neuron and MP\_LIF neuron using BackPropagation Through Time (BPTT) (\cite{lillicrap2019backpropagation}). In BPTT, the network is unrolled in the time dimension to calculate the weight update value. The details of the weight update process are as follows:

\begin{equation}
\label{eq:2}
\Delta w^{l}=\sum_{n} \frac{\partial \mathcal{L}_{\text {total }}}{\partial o_{t}^{l}} \frac{\partial o_{t}^{l}}{\partial V_{t}^{l}} \frac{\partial V_{t}^{l}}{\partial w^{l}}
\end{equation}

\begin{equation}
\label{eq:3}
\frac{\partial o_{t}^{l}}{\partial V_{t}^{l}}= \begin{cases}H_{1}^{\prime}\left(V_{t}-V_{t h}\right) & \text { if } o_{t}^{l}=S_{t}^{l} \\ 1 & \text { if } o_{t}^{l}=V_{t}^{l}\end{cases}
\end{equation}
where ${o_{t}^{l}}$ is the output of the neuron in time t, and ${V_{t}^{l}}$ is the membrane potential of LIF and MP\_LIF neuron. ${S_{t}^{l}}$ is the output spike of the LIF neuron in the time t. For LIF neuron, ${o_{t}^{l}}$ is not differentiable, BPTT utilizes surrogate gradient method to calculate it. The shifted Arctan function $H_1(x) = \frac{1}{\pi} arctan(\pi x)+ \frac{1}{2}$ is used as the surrogate function to replace the Heaviside function of LIF neuron. For MP\_LIF neuron, ${o_{t}^{l}} = {V_{t}^{l}}$ and $\frac{\partial o_{t}^{l}}{\partial V_{t}^{l}} = 1$, which is similar to an ANN activation function.

\section{Implementation And Results}
In this section, we focus on the implementation results of our proposed framework for retinal prostheses. Additionally, we discuss the event dataset performance, prediction performance of the SRNN model, and energy efficiency characteristic of the framework. Also, we compare the features of the proposed framework with other state-of-the-art frameworks.

\subsection{Event dataset performance}

\begin{figure}[!t]
\centering
\includegraphics[width= 0.5\textwidth]{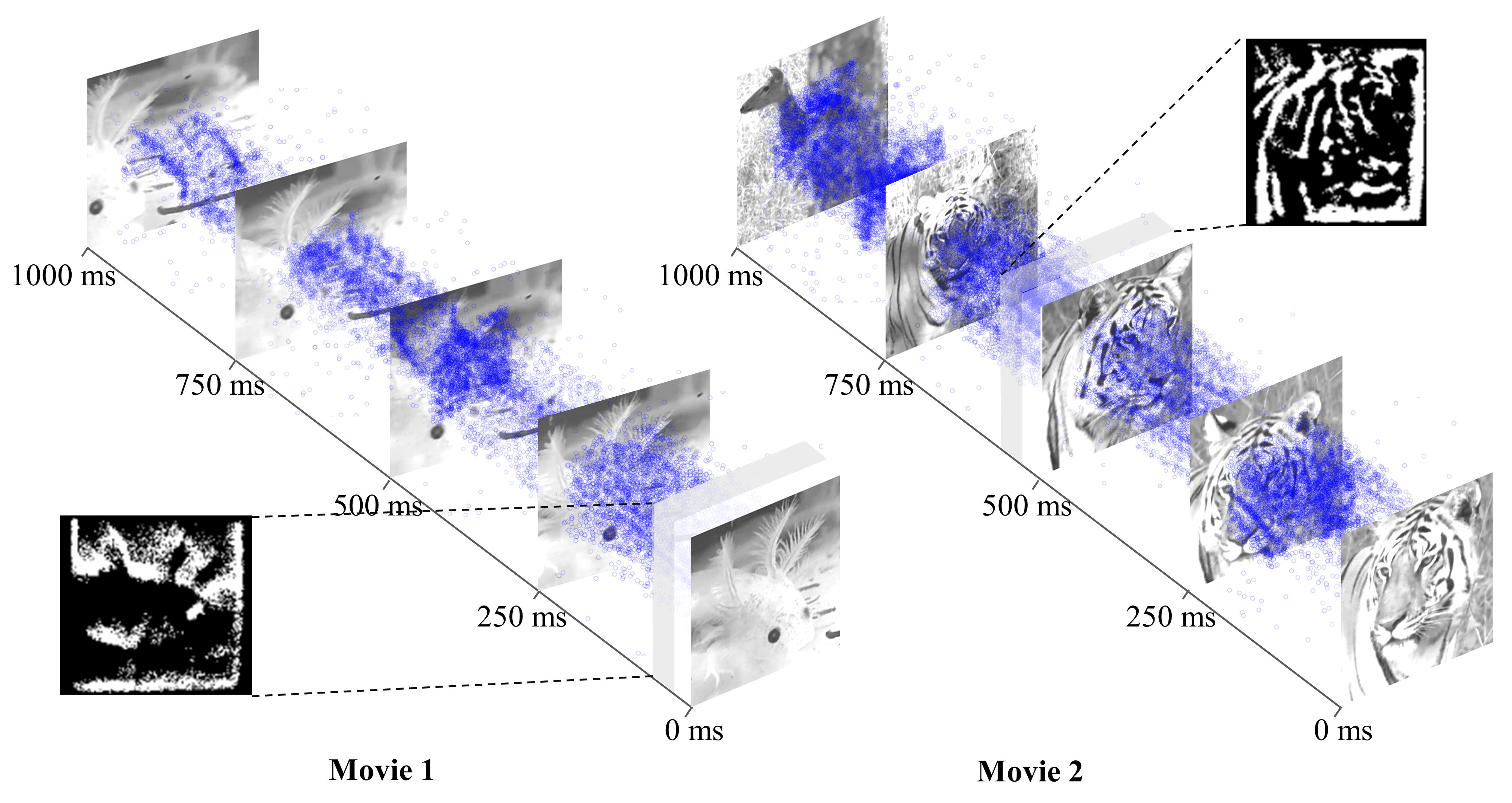}
\caption{Spike train and constructed binary pattern of two movies. The left figure shows the sparse spike train of a slow swimming salamander. And the binary image was built with the spike sequence in one time window, which includes shape and edge features. The right part demonstrates the spike train of the process of tigers preying on deer. The binary image proves the quality of the spike train could support the SRNN model to predict.}
\label{event_dataset}
\end{figure}

The recorded spike train of one second of two movies is depicted in Fig. \ref{event_dataset}. The original movie of the left figure displays a slowly swimming salamander, whose recorded spike train is sparse and consumes low energy in the perception process. In addition, the sparse spike train lessens the energy consumption during communication between the sensor and SRNN model. To verify the quality of the spike train, a binary image is built with the spike train in one time window. This image has a clear edge and shape information, providing efficient input signals for the SRNN model. The right figure shows the spike train of the other movie where a tiger preys on the deer. The dynamic vision sensor with a high temporal resolution still records the completed movement of the whole scene. Besides, it is easy to recognize a tiger through the constructed binary image. Thus, the quality of the recorded spike train could support the SRNN model to predict the response of ganglion cells.

\begin{figure}[!t]
\centering
\includegraphics[width= 0.5\textwidth]{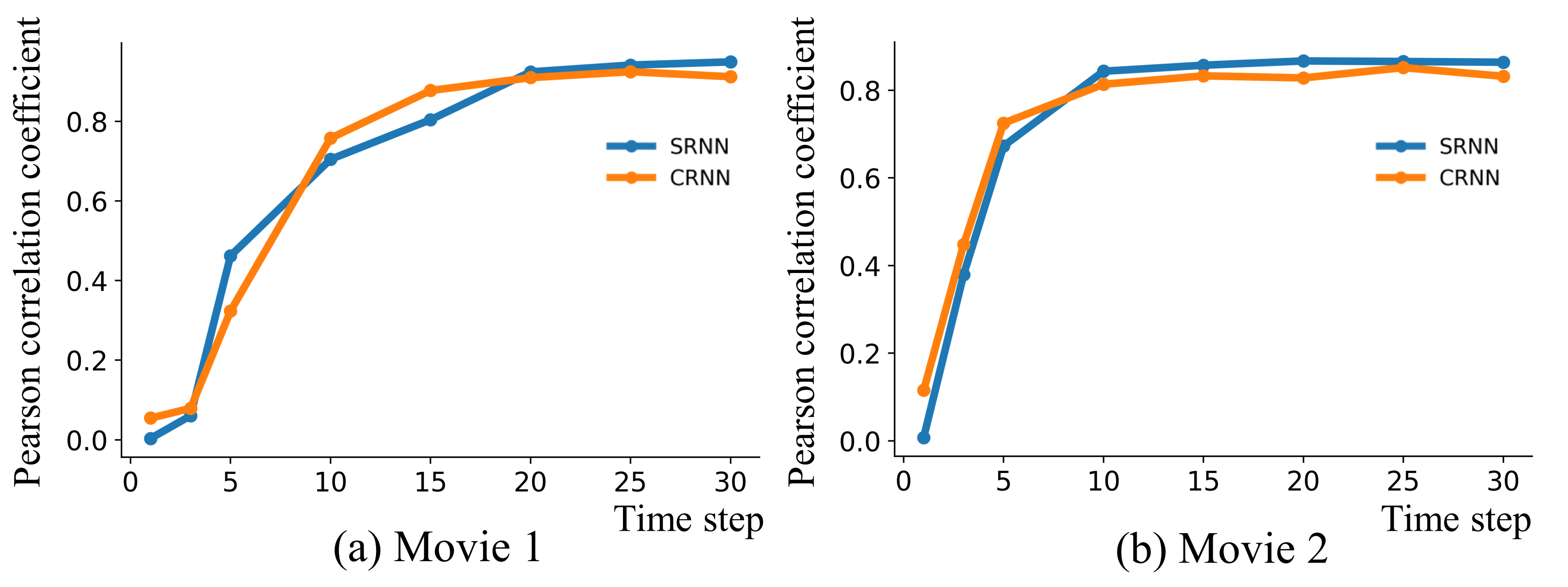}
\caption{Performance of SRNN and CRNN models is evaluated by the Pearson correlation coefficient between the recorded data and predicted value. The time step is an important factor in affecting the performance: (a,b) As the time step increases, the model performance on two movies increases as well. }
\label{time_step}
\end{figure}

\begin{figure*}[!t]
\centering
\includegraphics[width= 1.0\textwidth]{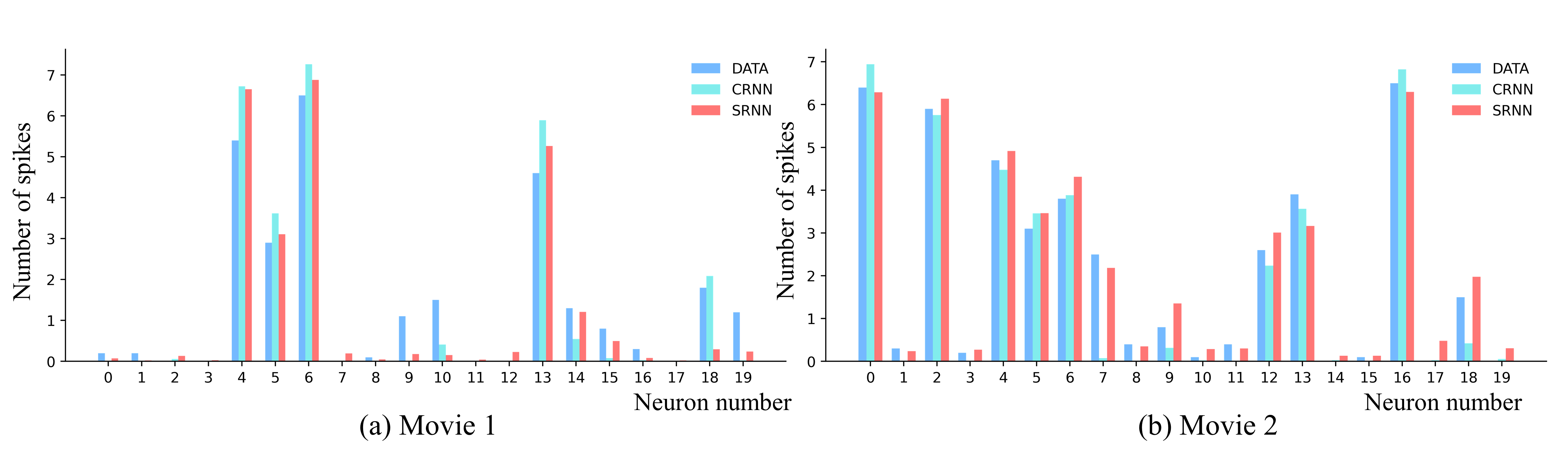}
\caption{The front 20 neurons were randomly selected from 38 neurons to show the performance. DATA in this figure represented the recorded number of spikes of ganglion cells: (a, b) These two histograms compared ganglion cells' responses with the predicted results of the CRNN and SRNN models in the two movies.
}
\label{spatial}
\end{figure*}

\begin{figure*}[!t]
\centering
\includegraphics[width= 1.0\textwidth]{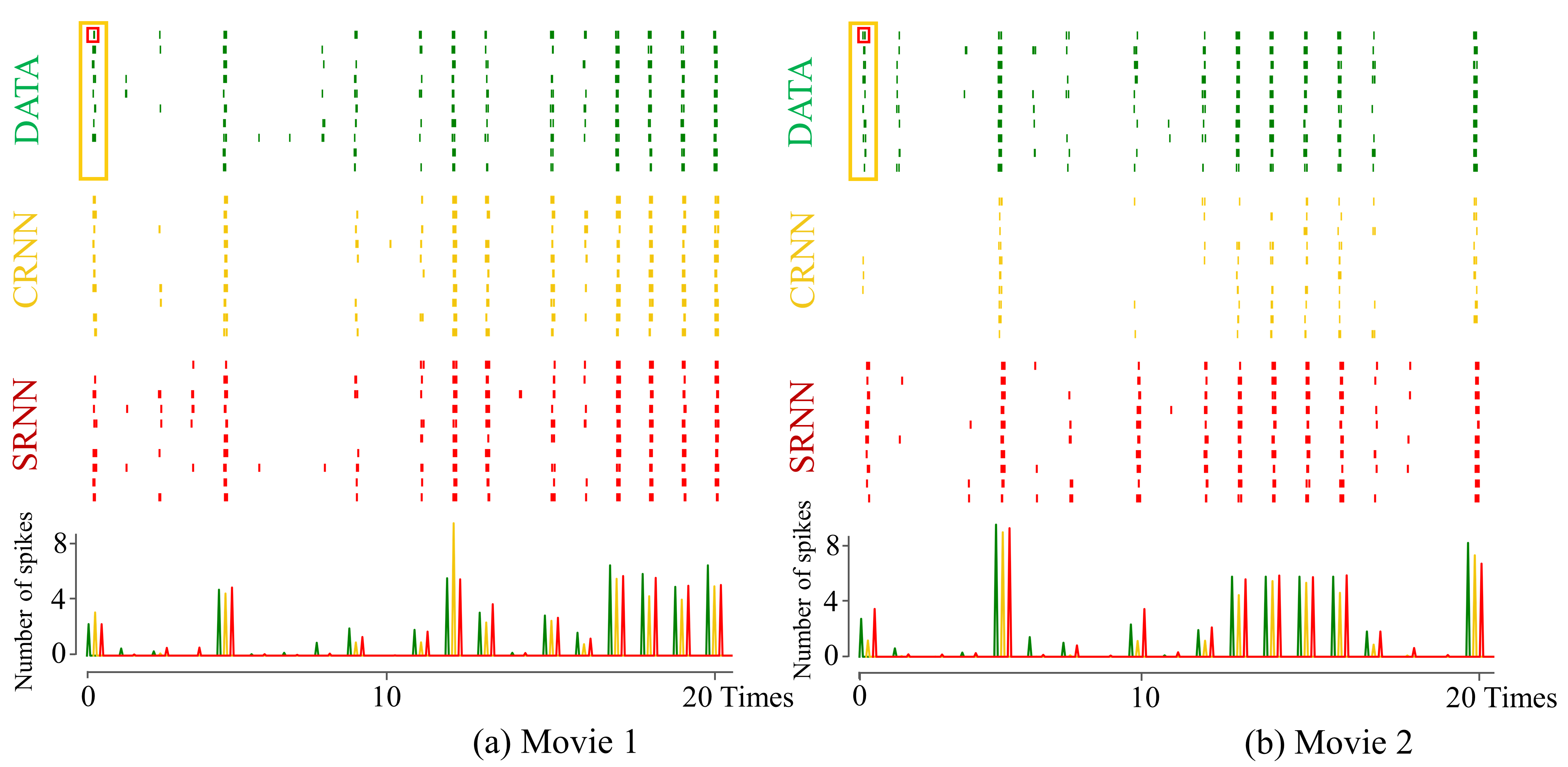} 
\caption{
Comparison between the spike train of the DATA, CRNN, and SRNN models of the front 20 times prediction of two movies. The spike train in the red rectangle was generated according to the corresponding firing rate of one neuron. And this spike train's time distribution and spike number followed the Poisson distribution. In addition, this generated process was repeated 10 times to avoid randomness, as shown in the yellow rectangle. The spike train of the CRNN and SRNN model adopted the same generated method. The comparison between the spike number of recorded data and the predicted values of CRNN and SRNN model of two movies were depicted in the bottom column. The comparison in spike train and spike number proved the predicted results of the SRNN model could fit well with the recorded data, outperforming the performance of the CRNN model.}
\label{temporal}
\end{figure*}

\subsection{Prediction performance of SRNN model}
The performance of the SRNN model is highly related to the time step parameter. Hence, we firstly explore the effect of time step on prediction performance. After selecting one specific time step, we compare the accuracy between the SRNN model and CRNN model in spatial and temporal domains. Finally, the deep learning model already has three common recurrent layers, including LSTM, GRU, and Vanilla RNN. Thus, it is necessary to compare the performance of the SRNN model with these three recurrent layers, which are used to replace the recurrent block in our proposed SRNN model. 

\textbf{1) SRNN model with different time steps: }
The performance of the two movies' SRNN and CRNN model of different time steps are depicted in Fig. \ref{time_step}. Their performance is evaluated with the Pearson correlation coefficient (PCC) between the predicted firing rate and recorded response. This figure demonstrates that as the time step increases, the models' PCC of two movies increases with more input signals. However, the large time steps can lead to a long delay for waiting the input signals and processing. One-time step represents the spike train of 33 ms, which means 30-time steps cause one-second delay. To balance the time delay and predicted accuracy, we selected 20-time steps as the experiment setup. In this time step, the PCC of the SRNN model outperforms the CRNN model in both used movies. And it achieves 0.93 and 0.86 in these two movies, respectively.

\begin{figure}[!t]
\centering
\includegraphics[width= 0.40\textwidth]{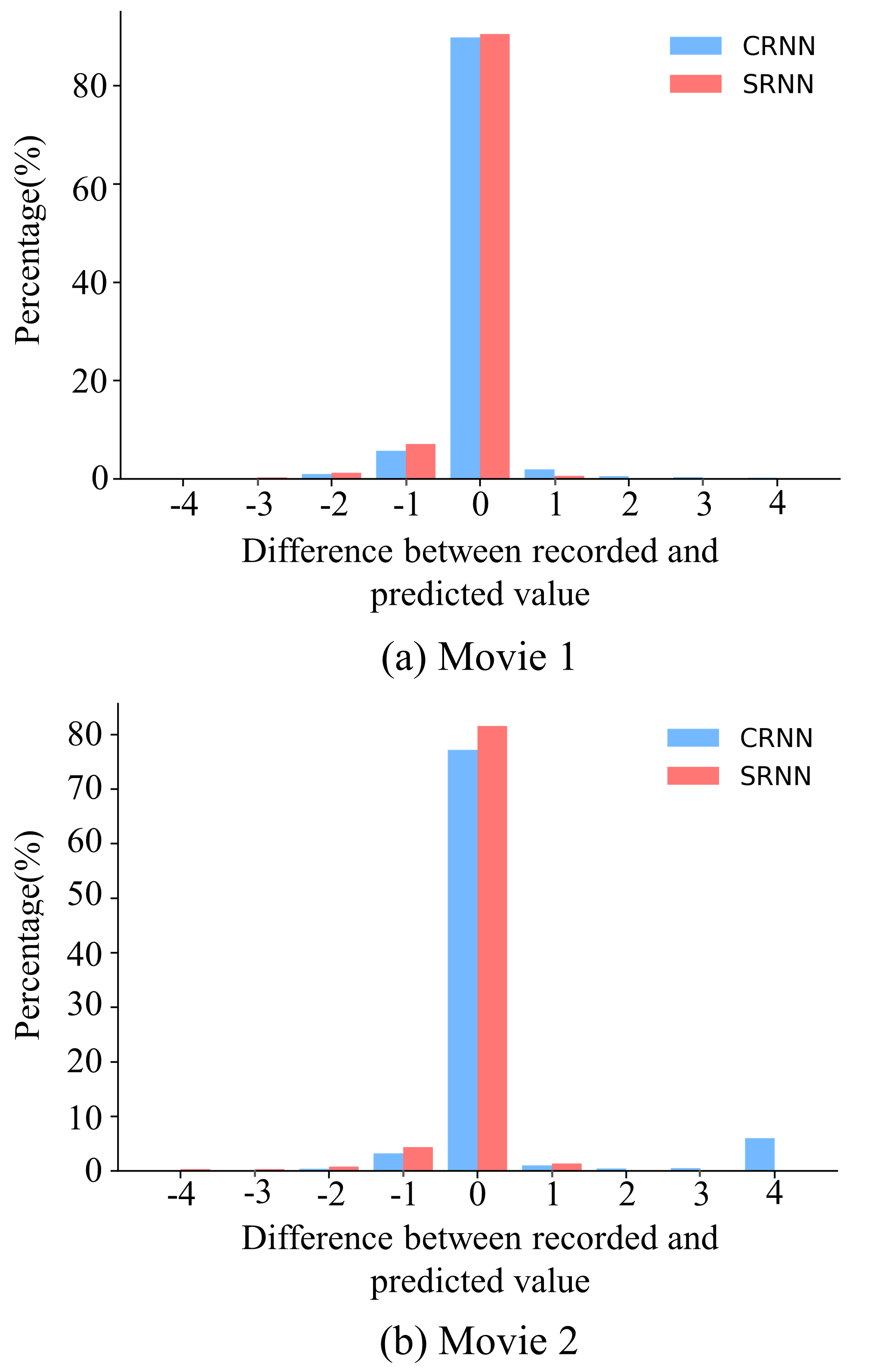}
\caption{The model performance in spatiotemporal domain was evaluated by the difference between recorded and predicted values in all neurons, which was obtained by subtracting the recorded firing rates from predicted values: (a) The difference distribution of the SRNN model is similar to the CRNN model in movie1, and the difference within [-0.5, 0.5] achieves 90\%, (b) The difference distribution of the SRNN model is better than the CRNN model in movie2, reaching 82\% in the range from -0,5 to 0.5.}
\label{spatiotemporal}
\end{figure}

\textbf{2) Comparison between SRNN model and CRNN model: }
We further compare the performance of the SRNN and CRNN models in spatial, temporal, and spatiotemporal domains. In the spatial domain, 20 neurons are randomly selected from the recorded 38 neurons to show the comparison results. DATA in Fig. \ref{spatial}(a) represents the recorded firing rate of ganglion cells. The predicted values of the SRNN model fit well with the recorded data for neurons with a high-firing rate, which is better than the CRNN model. However, both models have limited performance in neurons with a low-firing rate. In Fig. \ref{spatial}(b), the predicted value of the SRNN model is similar to the recorded data in almost all neurons, including high and low firing rate neurons. While the CRNN model still has a high similarity with neurons with a high-firing rate. It performs poorly in fitting the response of neurons with a low-firing rate. 

To show its performance in the temporal domain, we randomly selected one neuron from recorded 38 neurons. The comparison result of the first 20 times prediction of two movies is shown in Fig. \ref{temporal}. The first column in Fig. \ref{temporal}(a) shows the generated spike trains according to the recorded data. The spike train in the red rectangle displays the spike number and time distribution of one neuron in 660 ms, corresponding to 20-time steps. And this generated process follows Poisson distribution, repeated 10 times to avoid randomness. The spike train of both CRNN and SRNN models adopts the same method. The comparison results show that the spike train of the SRNN model can fit well with the recorded data for high-firing rate neurons, slightly better than the CRNN model. Both models have limited performance on low-firing rate neurons. The bottom column demonstrates the spike number comparison, which proves the high similarity between the recorded data and the predicted values of the SRNN model. 
In Fig. \ref{temporal}(b), the comparison results in spike train and bottom spike number prove the SRNN model has a good performance in fitting the response of all neurons to movie 2. In comparison, the CRNN model has poor performance in modeling the spike number of neurons with a low-firing rate. 

The similarity in the spatiotemporal domain is evaluated by the difference between recorded data and predicted value in all neurons. For movie 1, the performance of the SRNN model is similar to the CRNN model (Fig. \ref{spatiotemporal}(a)). The 0 in the x-axis represents the difference between the data and the model’s value belongs to [-0.5, 0.5], which achieves 90\% in this movie. The performance of the SRNN model is better than the CRNN model in movie 2, and the difference within [-0.5, 0.5] achieves 82\% (Fig. \ref{spatiotemporal}(b)). The spatial, temporal, and spatiotemporal comparison proves that the predicted spike number of the SRNN model can fit well with the recorded data, outperforming the CRNN model's performance.

\textbf{3) SRNN model with different recurrent layers: }
\begin{figure}[!t]
\centering
\includegraphics[width= 0.43\textwidth]{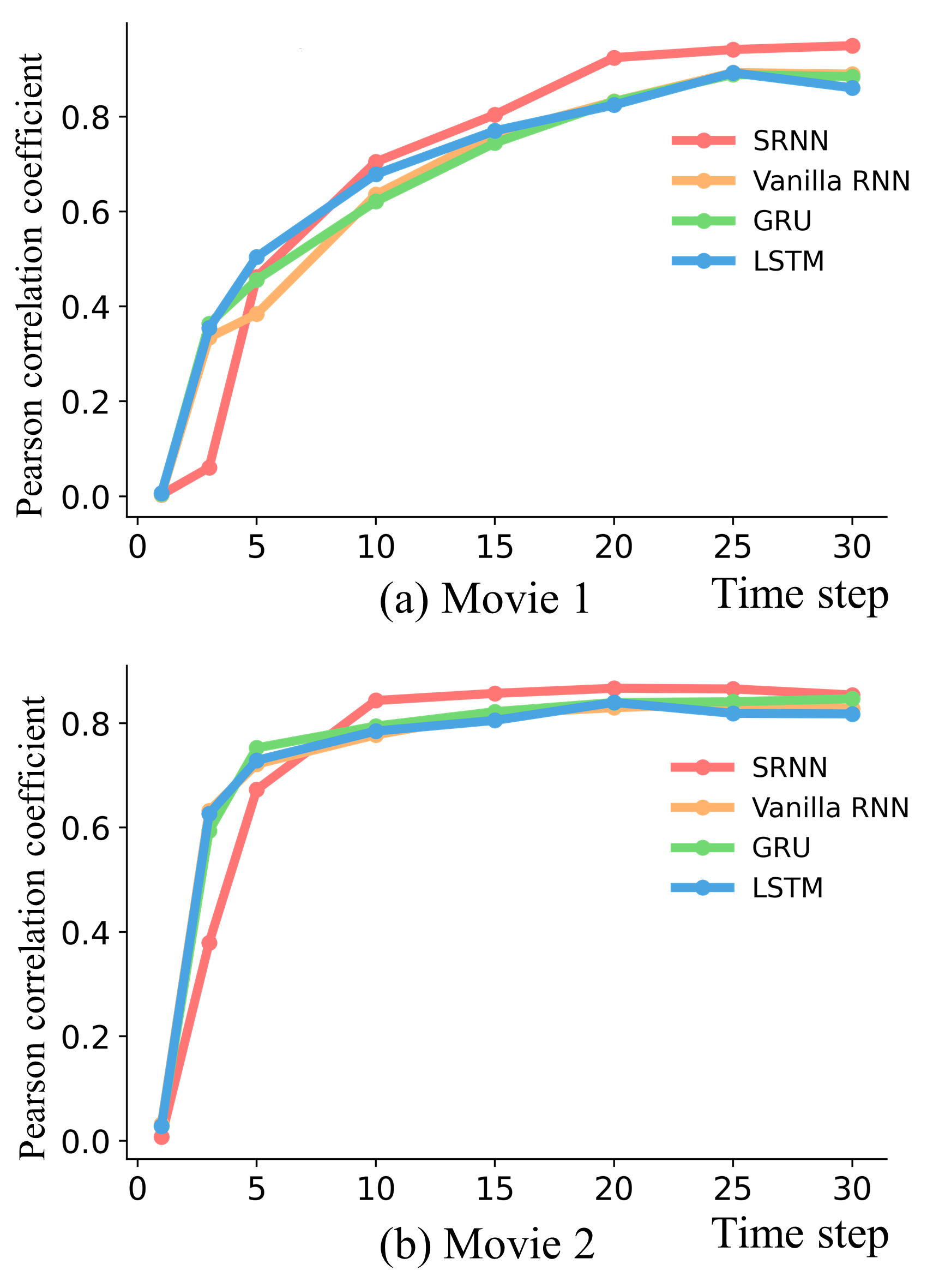}
\caption{The performance of three traditional recurrent layers is used to replace recurrent blocks in our proposed SRNN model: These three recurrent layers include Vanilla RNN, GRU, and LSTM, (a, b) The PCC of these layers is better than our proposed model when the time step is lower than 10. However, the performance of the SRNN model becomes better as the time step increases.
}
\label{recurrent_layer}
\end{figure}

In our proposed SRNN model, two recurrent blocks are adopted to make computation efficient and diversiform. However, there are already three different recurrent layers in the deep learning field: Vanilla RNN, GRU, and LSTM. Thus, it is necessary to compare the performance of the SRNN model with these traditional recurrent layers that are used to replace recurrent blocks in our proposed SRNN model. As is shown in Fig. \ref{recurrent_layer}(a, b), the PCC of three different recurrent layers is similar for the two movies. The results demonstrate the recurrent layer as a general form, rather than the specific structure of these layers, plays an important role in predicting the response of ganglion cells. In addition, when the time step is lower than 10 in movies 1 and 2, its performance exceeds our proposed SRNN model. But for the high time step, the PCC of the proposed SRNN model is better than these traditional recurrent layers.

\subsection{Energy efficient}

\begin{table}
\caption{Comparison of power consumption (per second) between the CNN processing-based framework and proposed NeuroSEE}
\centering
\begin{tabular}{c  c  c  c } 
\hline
Layer & \makecell{Spiking Neuron\\ Number} & \makecell{Neuron\\ Type} & \makecell{Spike Firing\\ Rate} \\
\hline
\makecell{First spike\\ layer} & 100352 & LIF & 5.29\% \\

\makecell{Second spike\\ layer} & 32768 & LIF & 5.94\% \\

\makecell{Third spike\\ layer} & 128 & LIF &  27.07\% \\
\hline
\multicolumn{3}{c}{Overall spike firing rate} &  5.47\% \\
\hline
\end{tabular}
\label{firing_rate}
\begin{tablenotes}
    \footnotesize
    \item[1] LIF = Leaky integrate and fire neuron.
\end{tablenotes}
\end{table}

\begin{table}
\caption{Comparison of power consumption (per prediction) between the CRNN processing-based framework and proposed SpikeSEE}
\centering
\begin{tabular}{|c | c | c | c | c | c | } 
\hline
 & \multicolumn{2}{c|}{Sensor} & \multicolumn{2}{c|}{Neural Network} & \multicolumn{1}{c|}{Total} \\
\hline
\multirowcell{6}{CRNN\\processing\\framework}& \multirowcell{6}{CIS} & \multirowcell{6}{95.04\\ mJ}  & C1 & 5.77 mJ  & \multirowcell{6}{103.87\\ mJ}\\
\cline{4-5}
 &  &  &C2   &3.01 mJ &   \\
\cline{4-5}
 &  &  &LSTM   &41.02 $\mu$J &   \\
\cline{4-5}
 &  &  &Fc   &11.18 nJ &   \\
\cline{4-5}
 &  &  &\makecell{CRNN\\model} & 8.83 mJ  &   \\
\hline
\multirowcell{8}{SpikeSEE:\\spike-based\\processing\\framework}& \multirowcell{8}{DVS} & \multirowcell{8}{6.6\\ mJ}  & S1 & 1.13 mJ  & \multirowcell{8}{8.33\\ mJ}\\
\cline{4-5}
 &  &  &S2   &590 $\mu$J &   \\
\cline{4-5}
 &  &  &S3   &2.3 $\mu$J &   \\
\cline{4-5}
 &  &  &R1   &2.07 $\mu$J &   \\
 \cline{4-5}
 &  &  &R2   &2.3 $\mu$J &   \\
\cline{4-5}
 &  &  & Rd  &23.34 nJ &   \\
\cline{4-5}
 &  &  &\makecell{SNN\\model}    & 1.73 mJ  &   \\
\hline
\end{tabular}
\label{comparison_power_consumption}
\begin{tablenotes}
    \footnotesize
    \item[1] C1,2,3 = Convolution layer 1,2,3; Fc = Fully connection layer;
    \item[2] S1,2,3 = Spike layer 1,2,3; R1,2 = Recurrent block 1,2; 
    \item[3] Rd = Readout layer; Total = sensor + neural network.
\end{tablenotes}
\end{table}

\textbf{1) Spike firing rate: }
The low-power consumption characteristic of Spiking neural network depends on the low spike firing rate. Our proposed SRNN model adopts three spike layers to extract features. Thus, we summarize the spike firing rate of these spike layers in Table \ref{firing_rate}. The first spike layer has 100352 LIF neurons, and its firing rate is 5.29\%. And the second and third layers have 32768 and 128 neurons, respectively. The corresponding firing rate is 5.94\% and 27.07\%, and the overall spike firing rate is 5.47\%. The low firing rate could contribute a lot to decreasing the power consumption, especially computing in an application-specific integrated circuit (ASIC) that can skip the calculation of non-firing LIF neurons. 

\textbf{2) Energy comparison: }
The energy-efficient feature is an important factor in the processing framework of retinal prostheses that needs to be used for a prolonged time. Thus, we compare the power consumption of the proposed SpikeSEE with a state-of-the-art CRNN processing-based framework. SpikeSEE adopts a dynamic vision sensor (DVS) to perceive external scenes and the SRNN model to fit the response of ganglion cells. CRNN processing-based framework utilizes CMOS image sensor (CIS) to record scene information and CRNN model to predict. 

The comparison of power consumption between SpikeSEE and CRNN processing-based framework is summarized in Table \ref{comparison_power_consumption}. The processing model in SpikeSEE adopts 20 time steps in one prediction, whose duration is 660 ms. Thus, we compare the energy consumption of two processing frameworks in 660 ms. For CRNN processing-based framework, the energy is composed of a CMOS image sensor and a CRNN model. In this experiment, TI PYTHON 300 was selected for CRNN processing-based framework since its resolution is similar to Inivation-DAVIS346 adopted in SpikeSEE (\cite{berner2013240, Wu2019HighPerformanceCI}). Because the power of this device is 80 mA@1.8 V DC, the energy of one prediction is 95.04 mJ. The energy of the CRNN model on a hardware platform is estimated by one power consumption estimation method, where the multiplication and addition energy of 32-bit floating-point operation are 3.7 and 0.9 pJ, respectively (\cite{horowitz20141}). Thus, the energy of the CRNN model in one prediction is 8.83 mJ. And the total power consumption of the CRNN processing-based framework in one prediction is 103.87 mJ. For SpikeSEE, the energy of dynamic vision sensor to record spike train within 20 time steps needs 6.6 mJ. In addition, the energy of the SRNN model to predict is 1.73 mJ. The total power consumption of the proposed SpikeSEE is 8.33 mJ, which is lower than CRNN processing-based framework and much more suitable for retinal prostheses.

\subsection{Comparison with state of the art works}

The comparison of various features between proposed SpikeSEE and related works is summarized in Table \ref{related_work}. The framework presented in (\cite{mcfarland2013inferring}) utilizes a traditional Linear-Nonlinear model to mimic the behavior of one ganglion cell. Its weak feature extraction ability restricts fitting performance at a low level. Another framework proposed a conductance-based neural encoding model to enhance the feature processing ability (\cite{latimer2019inferring}). In addition, a CNN processing-based framework utilizes the CNN model to extract more spatial features for predicting; its performance achieves 0.7 with Pearson correlation coefficient as evaluation standard (\cite{yan2020revealing}). However, these two frameworks can only model the response of one ganglion cell, which can not meet the requirement of retinal prostheses. Then, a framework presented in (\cite{zheng2021unraveling}) adopts the CRNN model to improve the extraction ability on the temporal domain. It achieves state-of-the-art performance and can model the response of multiple ganglion cells simultaneously. The CRNN model processes input signals with floating point multiplication, causing high-power consumption. Our proposed SpikeSEE utilizes DVS to encode video signals and the SRNN model to predict. The spike representation encoding technique of DVS and spike processing method of SRNN model can fit well with the response of ganglion cells in low-power conditions.

\begin{table*}
\caption{Comparison of main features between our proposed SpikeSEE and related works}
\centering
\begin{tabular}{c c c c c c} 
\hline
\textbf{Ref} &  \makecell{McFarland et al.\\ (2013)} & \makecell{Latimer et al.\\ (2019)} & \makecell{Yan et al.\\ (2020)} & \makecell{Zheng et al.\\ (2021)} & \textbf{This work}\\
\hline
\textbf{Bio-inspired} & \makecell{Inspired by the \\ receptive field\\ characteristic}  & \makecell{Inspired by the \\conductance transmission\\ mechanism} & No & No & \textbf{\makecell{Inspired by the \\ spike-based processing \\ and special structure\\ of human retina}} \\
\hline
\textbf{Perception} &  CIS & CIS & CIS & CIS & \textbf{DVS}\\
\hline
\makecell{\textbf{Temporal resolution}}  & 16 ms & 0.1 ms & 33 ms & 33 ms & \textbf{1 $\mu$s} \\
\hline
\textbf{Processing} & \makecell{Linear-Nonlinear\\ model}  & \makecell{Conductance-based\\ neural encoding\\ model} & CNN model & CRNN model & \textbf{SRNN model}\\
\hline
\textbf{Targeted cell} & Ganglion cell  & Ganglion cell & Ganglion cell & Ganglion cells & \textbf{Ganglion cells}\\
\hline
\makecell{\textbf{Predict the NoN}} & 1 & 1 & 1 & 80 & \textbf{38} \\
\hline
\textbf{Performance} & PCC: 0.4 & ;\makecell{The variance\\ of PSTH: 0.86} & PCC: 0.7 & PCC: 0.92 & \textbf{PCC: 0.93} \\
\hline
\makecell{\textbf{Power consumption}}  & N/A & N/A & N/A & 103.87 mJ & \textbf{8.33 mJ} \\
\hline
\end{tabular}
\label{related_work}
\begin{tablenotes}
    \footnotesize
    \item[1] NoN = Number of neurons; N/A = None applicable; PCC: Pearson correlation coefficient; 
    \item[2] PSTH = Peri-stimulus time histogram;
\end{tablenotes}
\end{table*}

\section{Discussions}

As mentioned in the section \uppercase\expandafter{\romannumeral1}, the prediction performance and power consumption are two important factors of the processing framework for retinal prostheses. The SRNN model of our proposed SpikeSEE improves the prediction accuracy with a bio-inspired structure, outperforming the state-of-the-art CRNN model. In addition, the spike representation encoding technique of the dynamic vision sensor and the spike processing method of the SRNN model significantly decrease the power consumption, making this framework more suitable for retinal prostheses. 

There is one unnoticed issue related to the processing performance of the framework, which is the time step parameter. The results section summarizes the effect of the time step on prediction accuracy. And a 20-time step was adopted in the framework to balance the accuracy and the processing delay. Because the input signal of one-time step is a spike train of 33 ms, 20 time steps mean a delay of 660 ms. The long delay of the framework is not suitable for the actual use of retinal prostheses. 

To decrease processing delay, an in-depth study of the human retina working mechanism and improving the quality of nerve signal acquisition are two potential solutions (\cite{baden2020understanding, yu2020toward}). Nowadays, human beings have a limited understanding of how the retina works and cannot fully understand how it encodes information and processes it (\cite{gollisch2010eye}). This research status restricts the performance of a biophysical processing-based framework that could only fit the stimuli and neural response of one ganglion cell. In addition, some bio-inspired deep learning processing-based frameworks have also shown performance bottlenecks (\cite{yan2020revealing, safarani2021towards}). Thus, neuroscientists need to further explore the perception and processing mechanisms of the retina, providing computational scientists with more biological basis and inspiration to design better algorithm models. In addition, improving the quality of neural signal acquisition is another possible way to reduce processing delay (\cite{wang2022neurosee}). The input data with low-time steps is difficult to establish the mapping relationship with the acquired neural signal.
This is most likely since the recorded animal did not focus on the video when the signal was collected. Therefore, designing a better experimental protocol to collect neuronal response data is an idea worth verifying.

The existing problem and potential solutions for prediction performance and power consumption at the software level have been fully discussed. It is also necessary to explore these two issues from the hardware level before applying the framework to retinal prostheses. Application-specific integrated circuit (ASIC) is a suitable hardware platform to implement the algorithm. This custom integrated circuit (IC) should support parallel computing on multiple levels to decrease the compute latency (\cite{chen2016eyeriss}). In addition, the low firing rate of the proposed SRNN model also needs to be fully used by the designed IC to reduce the calculation and energy consumption (\cite{kuang202164k}). A fully functional and powerful IC can solve the above problems, which is the core component of retinal prostheses.

A completed retinal prosthesis is composed of a sensor, a processing circuit, and a photostimulator (\cite{montazeri2019optogenetic}). The power consumption of each component should be minimized during the design process. The energy of communication between different components also needs to be reduced. In addition, the stimulation performance of the photostimulator is highly related to the final perception of the visual cortex. Thus, the mapping relationship between stimulation patterns of photostimulator and the response of ganglion cells should be established. Finally, a powerful and energy-efficient retinal prosthesis could bring versatile devices to facilitate the enhancement of the eyes' functions.

\section{Conclusion}
We proposed an energy-efficient dynamic scenes processing framework (SpikeSEE) to perceive video signals and fit the response of ganglion cells. The framework utilizes a spike representation encoding method to encode dynamic videos into sparse spike trains, which can provide rich edge and shape features for subsequent processing models. In addition, a bio-inspired SRNN model is used to predict the firing rate of ganglion cells to video signals. Its predicted results are similar to the recorded response in spatial and temporal domains, outperforming the performance of the state-of-the-art CRNN model. Additionally, this framework with spike-based encoding and processing methods requires only (1/12)th of the power consumption of the CRNN processing-based framework. In summary, the proposed SpikeSEE with higher prediction accuracy and low-power consumption is a better choice to implement mimic retinal prostheses. 

\section{Acknowledgments}
The authors acknowledge start-up funds from Westlake University to the Center of Excellence in Biomedical Research on Advanced Integrated-on-chips Neurotechnologies (CenBRAIN Neurotech) for supporting this research project. This work was funded in part by the Zhejiang Key R\&D Program Project No. 2021C03002, and in part by the Zhejiang Leading Innovative and Entrepreneur Team Introduction Program No. 2020R01005.








\bibliographystyle{cas-model2-names}
\bibliography{my_ref}



\end{document}